\documentclass[review]{elsarticle}

\usepackage{graphicx}
\usepackage{amsmath}
\usepackage{amssymb}
\usepackage{booktabs}
\usepackage{microtype}
\usepackage{pifont}
\usepackage{enumitem}
\usepackage{ragged2e}
\usepackage{xcolor}
\usepackage{times}
\usepackage{subfigure}
\usepackage{url}

\usepackage{booktabs}
\usepackage{multirow}
\usepackage{pifont}
\usepackage{amsmath}
\usepackage{graphicx}
\usepackage{microtype}
\usepackage{adjustbox}
\usepackage{colortbl}
\usepackage{bm}
\definecolor{cvprblue}{rgb}{0.21,0.49,0.74}

\usepackage[pagebackref=false,breaklinks=true,letterpaper=true,colorlinks,bookmarks=false,citecolor=citecolor,linkcolor=linkcolor]{hyperref}

\definecolor{citecolor}{HTML}{0071BC}
\definecolor{linkcolor}{HTML}{ED1C24}

\journal{Pattern Recognition}

\begin{document}

\begin{frontmatter}

\title{MoPO: Incorporating Motion Prior for Occluded Human Mesh Recovery}

\author[pku]{Tao Tang}
\ead{taotang@stu.pku.edu.cn}
\author[pku]{Hong Liu\corref{cor}}
\ead{hongliu@pku.edu.cn}
\author[pku]{Xinshun Wang}
\ead{wangxinshun@stu.pku.edu.cn}
\author[pku]{Wanruo Zhang}
\ead{wanruo_zhang@stu.pku.edu.cn}

\cortext[cor]{Corresponding author: Hong Liu.}

\address[pku]{State Key Laboratory of General Artificial Intelligence, Peking University, Shenzhen Graduate School, China}
\begin{abstract}
Although recent studies have made remarkable progress in human mesh recovery, they still exhibit limited robustness to occlusions and often produce inaccurate poses and severe motion jitter due to the insufficient spatial features for occluded body parts. 
Inspired by the rapid advancements in human motion prediction, we discover that compared to occluded image features, pose sequence inherently contains reliable motion prior for estimating occluded body parts.
In this paper, we incorporate \textbf{Mo}tion \textbf{P}rior for \textbf{O}ccluded human mesh recovery, called \textbf{MoPO}. 
Our MoPO mainly consists of two components: 
1) The motion de-occlusion module, where we propose a spatial-temporal occlusion detector to detect joint visibility, and then we propose a lightweight motion predictor to complete the occluded body parts by predicting the most plausible joint positions based on history poses. 
2) The motion-aware fusion and refinement module, which fuses the completed joint sequence with image features to estimate human shape and initial human pose. 
Moreover, the completed joint sequence is further used to refine the final human pose through inverse kinematics, which provides the occlusion-free motion prior for regressing human poses. 
Extensive experiments demonstrate that MoPO achieves state-of-the-art performance on both occlusion-specific and standard benchmarks, significantly enhancing the accuracy and temporal consistency of occluded human mesh recovery. \textcolor{cvprblue}{Our code and demo can be found in the supplementary material}. 
\end{abstract}


\end{frontmatter}

\section{Introduction}
Recovering human meshes from monocular images is a crucial yet challenging task in computer vision, with broad applications in animation, virtual reality, AI coach, and robotics~\cite{meshsurvey}. Unlike 3D Human Pose Estimation (HPE)~\cite{graphmlp,posesurvey}, which estimates simple human movements through sparse skeletal joints, 3D Human Mesh Recovery (HMR)~\cite{meshsurvey} is a more complex task that aims to estimate the detailed 3D coordinates of human meshes. Based on the use of the parametric human model, existing HMR methods can be divided into model-based and model-free approaches.

Model-based methods~\cite{vibe, pare, jotr, dpmesh} aim to reconstruct human pose and shape by estimating the parameters of the human model (e.g., Skinned Multi-Person Linear (SMPL)~\cite{smpl}). 
These methods simplify the HMR task into estimating a small set of parameters instead of a large number of vertices.
Since the parameter space is trained on large-scale human datasets, the parametric human model provides strong priors (e.g. reasonable pose and shape) for HMR.
However, they tend to be less flexible due to the limited parameter space. 
On the other hand, model-free methods~\cite{pose2mesh, pmce}, regress vertex positions without any constraints on the parameter space, allowing them to capture local details of the human surface. 
However, due to the lack of human priors, these methods often result in unrealistic human poses and shapes in challenging scenarios.

Most existing HMR methods require the human body to be fully visible in images. 
Reconstructing an accurate human mesh from occlusion scenarios (self-occlusion, object-person occlusion, and person-person occlusion) remains a great challenge. 
Although occluded HMR methods~\cite{romp, pare, 3dcrowdnet, visdb, jotr, sefd, instancehmr, dpmesh} have made remarkable progress,
most of them rely on 2D human evidence (e.g., part segmentation masks~\cite{pare}, 2D keypoints~\cite{dpmesh} or edge maps~\cite{sefd}) to improve the alignment of human mesh with the visible body parts.
Additionally, some methods~\cite{jotr, bev} further introduce 3D representations to locate 3D joints and extract 2D features from corresponding image regions. 
Despite these efforts, reconstructing humans from occluded images remains an ill-posed problem due to the inherent difficulty of reasoning occluded body parts from insufficient image features. 
Furthermore, when existing occluded HMR methods are applied to videos, they often neglect the temporal consistency between frames, leading to severe motion jitter. 
As shown in Fig.~\ref{fig1}, When applied to real-world videos, existing occluded HMR methods often suffer from the following challenges:

\begin{figure}[t]
    \centering    
    \includegraphics[width=0.8\linewidth]{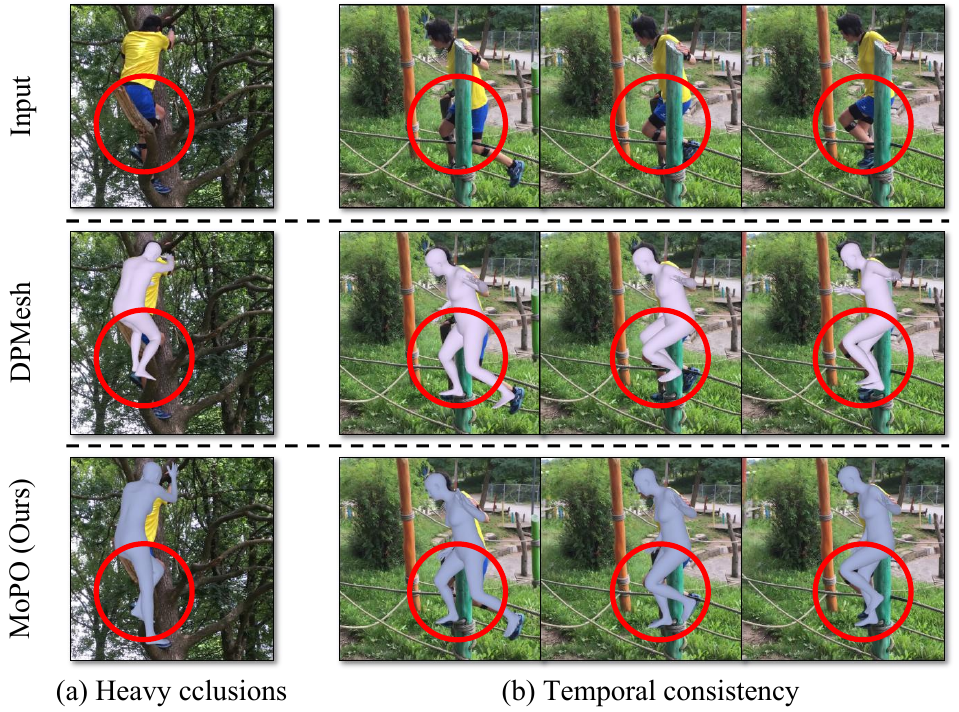}
    \vspace{-0.5em}
    \caption{
        \textbf{Comparisons between proposed MoPO and SOTA method.} Existing state-of-the-art occluded HMR method~\cite{dpmesh} suffers from (a) heavy occlusions and (b) temporal inconsistency. Our MoPO can produce accurate and temporally consistent pose and shape under diverse occlusions by incorporating motion prior.
    }
    \label{fig1}
    \vspace{-0.5em}
\end{figure}

\textbf{(i) Inaccurate human poses under heavy occlusions.} 
Heavy occlusions pose a significant challenge in accurately estimating the positions of occluded body parts.
The lack of spatial cues limits the network's ability to learn the highly non-linear mapping from limited image features to SMPL pose parameters~\cite{shapeboost}. Consequently, these methods often result in inaccurate and unnatural human poses.


\textbf{(ii) Severe motion jitter due to temporal inconsistency.} 
Existing occluded HMR methods typically rely on visible 2D or 3D human evidence(e.g., keypoints~\cite{3dcrowdnet}, human edges~\cite{sefd}) to reason occluded body parts. 
However, when applied to videos, the unstable estimation for each frame leads to severe motion jitter~\cite{mpsnet} under occlusions. 

Inspired by the rapid advancements in Human Motion Prediction (HMP)~\cite{hmpsurvey}, it is possible to predict the positions of occluded joints based on the motion prior from visible joint sequence. 
Subsequently, the completed joint sequence also provides a reliable motion prior for the parametric human model and reduces the motion jitter caused by occlusions.
Although DPMesh~\cite{dpmesh} also introduces priors from the text-to-image diffusion model~\cite{diffusion} to assist reasoning occluded body, its prior is trained from general scenes and has limited benefit under heavy occlusions (Fig.~\ref{fig1} (a)). Moreover, the implicit prior exhibits unstable predictions for similar human occlusions and causes temporal inconsistency (Fig.~\ref{fig1} (b)).

Based on the observations above, we incorporate \textbf{Mo}tion \textbf{P}rior into \textbf{O}ccluded HMR, called \textbf{MoPO}. 
As shown in Fig.~\ref{fig1}, our method enhances the robustness and temporal consistency of HMR under occlusions. 
Our MoPO addresses the occluded HMR in two stages:
1) Motion de-occlusion stage. First, occluded joints are detected based on the spatial confidence from 2D pose detector~\cite{vitpose, distribution} and temporal information from past frames. 
For occluded joints, we propose a lightweight MLP-based motion predictor to complete them based on the past joints, which provides occlusion-free and reliable motion prior for HMR.
2) Motion-aware fusion and refinement Stage. The completed joint sequence is extracted as motion context, which is further fused with image features~\cite{4dhmr, cliff, spin} to estimate SMPL shape and initial pose. 
Finally, we employ inverse kinematics to refine SMPL pose parameters directly from the completed joint sequence, alleviating the difficulties of regressing SMPL pose parameters with 6D representations~\cite{6d} under occlusions.

Our contributions are summarized as follows:
\begin{itemize}
\item We are the first to introduce motion complement-based de-occlusion for occluded human mesh recovery, allowing the explicit complement of occluded joints from the past joint sequence. The completed poses provide occlusion-free motion prior for robust human body estimation.
\item The motion-aware fusion and refinement module integrates the completed poses with the image features and refines SMPL pose parameters based on inverse kinematics, effectively utilizing the reliable motion prior for HMR under occlusions. 
\item Our MoPO achieves state-of-the-art performance on both occlusion-specific and standard benchmarks~\cite{3dpw, 3doh, cmu}. In particular, on the object-person occlusion dataset 3DPW-OC~\cite{3doh}, we significantly reduce the MPJPE and MPVPE by 8.5\% and 12.0\%, respectively.
\end{itemize}

\section{Related Work}
Our MoPO is a model-based method for occluded human mesh recovery from video.
Therefore we review the related works that can be divided into pose-based HMR, video-based HMR, and approaches for handling occlusion.
\subsection{Pose-based Human Mesh Recovery}
Pose-based methods recover human mesh from 2D or 3D human pose, which can naturally align the human mesh with joint positions. 
However, the inherent ambiguity in lifting 2D images to 3D human model gives rise to the need for prior knowledge. 
Human Pose Estimation (HPE) serves as a critical precursor in this pipeline, where recent advances have significantly improved the accuracy of pose priors. For instance, GraphMLP\cite{graphmlp} utilizes a unified global-local-graphical architecture to incorporate the graph structure of the human body into an MLP-based framework. ASFnet\cite{asfnet} explicitly incorporates depth cues to filter noise through an adaptive network.
Pose priors provide plausible predictions of human pose, helping to restrict the human mesh within a reasonable distribution~\cite{pliks}. 
For instance, Pose2Mesh~\cite{pose2mesh} designs a multi-stage meshnet to upsample sparse 3D poses to human mesh. GTRS~\cite{gtrs} introduces a graph Transformer network to reconstruct human mesh from a 2D human pose. Recently, some methods~\cite{niki, hybrik} leverage inverse kinematics to estimate SMPL pose parameters from human poses.
Although these pose-based methods have achieved remarkable performance in accuracy, they tend to regress an average human shape due to the lack of shape information within sparse joints.

\subsection{Video-based Human Mesh Recovery}
Compared to pose-based HMR methods, video-based HMR methods simultaneously recover accurate and temporally consistent human pose and shape. 
However, directly regressing SMPL parameters from the image features of each frame often results in unsmooth human motion. 
The previous video-based methods mainly focus on designing temporal extraction and fusion networks to enhance temporal consistency. 
For instance, VIBE~\cite{vibe}, MEVA~\cite{meva}, and TCMR~\cite{tcmr} carefully design Gated Recurrent Units (GRUs) based temporal extraction networks, which are utilized to smooth human motion. 
However, they struggle to model long-term dependencies by using GRUs. 
Consequently, most methods utilize Transformer-based temporal networks. 
MEAD~\cite{mead} proposes a spatial and temporal Transformer to parallel model these two dependencies. 
GLoT~\cite{glot} proposes a global and local Transformer to decompose the modeling of long-term and short-term temporal correlations. 
Bi-CF~\cite{bicf} introduces a bi-level Transformer to model temporal dependencies in a video clip and among different clips.
UNSPAT \cite{unspat} proposes a spatialtemporal Transformer to incorporate both spatial and temporal information without compromising spatial information. 
PMCE~\cite{pmce} utilizes cross-attention to fuse image features with the skeletons to incorporate the 3D pose prior.
ARTS~\cite{arts} propose a semi-analytical regressor to incorporate human poses through analytics and cross-attention.
Despite the complicated design of these temporal networks, when facing occlusions they often produce severe motion jitter due to the insufficient image features and unstable skeleton estimation. 

\subsection{Occluded Human Mesh Recovery}
Handling occlusions is challenging yet crucial for human mesh recovery. 
The straightforward solution is relying on visibility cues and most occluded HMR methods~\cite{pymaf, ochmr, visdb, 3dcrowdnet, jotr, instancehmr, dpmesh} utilize visible body to align the reconstructed human mesh with 2D or 3D human observations.
For instance, PARE~\cite{pare} utilizes an attention mechanism to infer occluded regions by gathering useful information from adjacent visible parts. 
VisDB~\cite{visdb} first predicts the coordinates and visibility labels of mesh vertices and then uses them to regress SMPL parameters. 
3DCrowdNet~\cite{3dcrowdnet} fuses the 2D pose heatmap with image features to focus on the areas corresponding to the occluded body.
JOTR~\cite{jotr} proposes a novel 3D joint contrastive learning approach with joint-to-joint and joint-to-non-joint contrastive losses for better supervision.
InstanceHMR~\cite{instancehmr} tackles person-person occlusion by encoding identity information at joint positions and body center representations, effectively representing body part ownership.
DPMesh~\cite{dpmesh} first leverages the powerful prior from a pre-trained text-to-image diffusion model by using 2D keypoints as a condition.
MEGA~\cite{mega} introduces a masked generative autoencoding framework that learns to reconstruct missing human features from large-scale data, reformulating human mesh recovery as a feature completion task.
Nevertheless, these neighboring visual cues are unreliable and insufficient for estimating occluded parts under heavy occlusions. 
Therefore, a few methods~\cite{glamr, chomp, occ-aware} introduce temporal information to reason the occluded parts and GLAMR~\cite{glamr} is the most related work with us. 
GLAMR~\cite{glamr} proposes a motion infiller, which leverages the learned latent code of human dynamics to infill the occluded humans through a Transformer-based network.
CHOMP~\cite{chomp} employs non-occluded human data to learn a joint-level spatial-temporal motion prior for occluded humans with a self-supervised strategy.
PostoMETRO~\cite{posto} leverages inter-part semantic correlations to infer plausible poses for occluded limbs. 
Although they achieve temporal consistent results under partial and short-term occlusions, they struggle with severe and long-term occlusions since the extracted implicit motion prior is less stable.
In contrast, our MoPO utilizes visible body joints from past frames to complete the occluded parts, which can provide explicit motion prior for more accurate and temporally consistent HMR.

\section{Methodology}
\subsection{Overall Framework}
\begin{figure}[t]
\centering
\includegraphics[width=\textwidth]{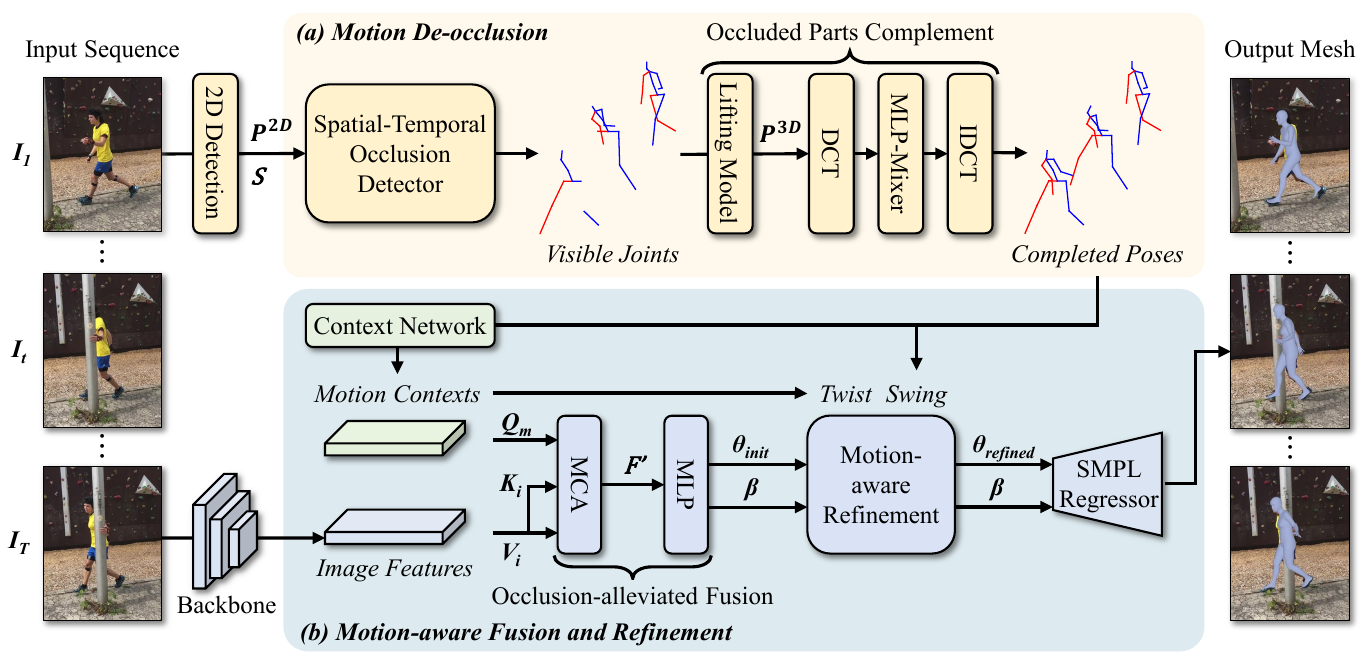}
\caption{ \textbf{Overview of the proposed MoPO.} Given a video sequence, a 2D detection is employed to obtain 2D human poses $\bm{P^{2D}}$ and confidence scores $\bm{S}$. Then, the motion de-occlusion module detects visible joints and completes occluded parts through an MLP-based motion predictor. 
Moreover, the motion-aware fusion and refinement module fuse the motion prior from completed poses $\bm{P^{3D\_com}}$ with image features to estimate SMPL shape $\bm{\beta}$ and init pose $\bm{\theta_{init}}$. The inverse kinematics is used to further refine the pose parameter $\bm{\theta_{refined}}$. Finally, MoPO feeds the refined SMPL parameters ($\bm{\theta_{refined}}$, $\bm{\beta}$) to the SMPL regressor to generate the human mesh.
}
\label{fig2}
\end{figure}

The overall framework of our occluded human mesh recovery method that incorporates motion prior is illustrated in Fig.~\ref{fig2}. 
MoPO mainly consists of two components: 1) motion de-occlusion and 2) motion-aware fusion and refinement. 
Specifically, given an image sequence $\bm{I}={\{I_t\}}^{T}_{t=1}$ with $T$ frames, we first employ a 2D pose detector~\cite{vitpose, distribution} to detect the 2D human poses $\bm{P^{2D}}\mathbb{\in R}^{T\times K\times 2}$ and confidence scores $\bm{S}\mathbb{\in R}^{T\times K\times 1}$ from images, where $K$ represents the number of keypoints. 
Meanwhile, a pre-trained backbone (ViT~\cite{4dhmr}) is used to extract the image features $\bm{F}\mathbb{\in R}^{T\times C_1}$ of each frame.
For the motion de-occlusion module, we first detect the occluded joints based on the spatial and temporal confidence of each joint.
Furthermore, we train a lifting network to lift the 2D poses to 3D poses $\bm{P^{3D}}\mathbb{\in R}^{T\times K\times 3}$ and iteratively complete the occluded human parts along the image sequence through a lightweight MLP-based motion predictor.
In the motion-aware fusion and refinement module, the completed poses $\bm{P^{3D\_com}}\mathbb{\in R}^{T\times K\times 3}$ is encoded into motion context. 
Then, we use cross-attention to fuse motion context with the image features for regressing SMPL parameters. 
Finally, we further optimize SMPL pose parameters through inverse kinematics and use the refined pose $\bm{\theta_{refined}}\mathbb{\in R}^{T\times 72}$ and shape $\bm{\beta}\mathbb{\in R}^{T\times 10}$ parameters to get the final human mesh sequence $\bm{M}\mathbb{\in R}^{T\times 6890\times 3}$.

\subsection{Motion De-occlusion}
\textbf{Spatial-temporal Occlusion Detector.}
Accurately distinguishing occluded human parts is critical for effective occluded human mesh recovery.
To achieve this, this module utilizes the confidence scores $\bm{S}$ generated from 2D joint detector~\cite{vitpose, distribution} to estimate the visibility of each joint. 
Unlike previous methods~\cite{dpmesh, 3dcrowdnet} that rely solely on the confidence scores from the current frame, we incorporate temporal information from past frames.
Specifically, for each joint in the $t$-{th} frame, we compute its confidence score as a weighted sum of the confidence from the $t$-{th} frame and $(t-1)$-{th} frame, which can be expressed as follows:
\begin{equation}
\bm{S'_{t}} = \alpha \times \bm{S_{t}} + (1 - \alpha) \times \bm{S_{t-1}},
\label{eq1}
\end{equation}
where $\bm{S'}$ denotes the spatial-temporal weighted confidence, $\alpha$ represents the weight.
In addition, the confidence of joints from $0$-{th} to $(t-1)$-{th} frames are processed by the RNN to estimate the confidence offset of the current frame, which effectively integrates the visible trend of previous frames. The equation is shown as follows:
\begin{equation}
\bm{S''_{t}} = \mathrm{RNN}(\{\bm{S_{t}}\}^{t-1}_{t=0}) + \bm{S'_{t}},
\label{eq2}
\end{equation}
where $\bm{S''}\mathbb{\in R}^{T\times K\times 1}$ represents the final confidence. Finally, joints across all frames with confidence scores above the threshold $thred$ are grouped into the visible joint set $V=\{v_t\}^T_{t=0}$, while those below $thred$ are grouped into the occluded joint set $OccJ=\{occj_t\}^T_{t=0}$. This module can detect a temporally consistent occluded joint set and reduce visibility jitter under occlusions.

\noindent\textbf{Occluded Parts Complement.}
Different from previous methods~\cite{3dcrowdnet, jotr} that focus on aligning the human mesh with the 2D or 3D human evidence from images, our MoPO leverages motion prior to complete occluded joints. 
Inspired by human motion prediction methods~\cite{hmpsurvey, simlpe}, we find that the occluded human parts can be accurately and consistently predicted from the visible joint sequence.
Therefore, we train an MLP-based motion predictor to iteratively complete occluded human parts in occluded joint set $OccJ$ from previous frames.
Specifically, we first design a lifting model to estimate 3D pose $\bm{P^{3D}}$ from 2D pose $\bm{P^{2D}}$ and then input the estimated 3D pose sequence from $0$-{th} to $N$-{th} frame to motion predictor. 
Then, the input pose positions are transformed into the frequency domain using a Discrete Cosine Transform (DCT) and we employ an MLP-Mixer~\cite{mlpmixer} Network to explore the spatial-temporal information within the pose sequence.
Finally, we predict the frequency representations for $N$-{th} to $(N+L)$-{th} frames, which are converted back to the pose positions through an Inverse DCT (IDCT). The detailed formula is as follows:
\begin{equation}
\bm{P^{3D}} = \mathrm{Lift}(\bm{P^{2D}}),
\label{eq3}
\end{equation}
\begin{equation}
\resizebox{.8\hsize}{!}{$\{\bm{P^{3D\_pred}_t}\}^{N+L}_{t=N}=IDCT(\mathrm{MLP\text{-}Mixer}(DCT(\{\bm{P^{3D}_t\}}^N_{t=0}))),$}
\label{eq4}
\end{equation}
where $N$ represents the input length of the motion predictor and $L$ denotes the predicted length, $\bm{P^{3D\_pred}}$ represents the predicted 3D pose.
If the detected pose for frame $N+1$ contains occluded joints set $occj_t$, we replace the occluded joints with the predicted joints. The formula is as follows:
\begin{equation}
  \bm{P^{3D\_com}_{N+1, k}} = 
  \begin{cases}
    \bm{P^{3D\_pred}_{N+1, k}}, & \text{if } k \in {occj}_{N+1}, \\
    \bm{P^{3D}_{N+1, k}}, & \text{if } k \in {v}_{N+1},
  \end{cases}
\end{equation}
where $k$ represents the $k$-{th} joints of human pose, $\bm{P^{3D\_com}}$ is the completed poses. 
After completing the $(N+1)$-{th} frame, we input $1$-{st} to $(N+1)$-{th} frames into the motion predictor to iteratively complete the entire occluded joint set.
Due to the lightweight of the spatial-temporal MLP-Mixer, the iterative complement is also efficient.
For segments with fewer than $N$ frames at the start or end of the sequence, we extend them by copying the initial frame. 

\subsection{Motion-aware Fusion and Refinement}
\textbf{Occlusion-alleviated Fusion.} The completed poses $\bm{P^{3D\_com}}$ provides an occlusion-free and reliable motion prior to alleviate the difficulty of occluded HMR. 
To fully utilize the motion prior, we first employ an RNN-based context network to extract motion features $\bm{F_{M}} \mathbb{\in R}^{T\times C_2} $ from the completed poses. 
Then, the flattened poses are concatenated with the motion features to form the motion context $\bm{C_{M}} \mathbb{\in R}^{T\times (C_2 + K\times 3)}$, as shown in the following equations:
\begin{equation}
 \bm{F_{M}} = \mathrm{RNN}(\bm{P^{3D\_com}}),
\label{eq6}
\end{equation}
\begin{equation}
 \bm{C_{M}} = concat[\bm{F_{M}}, Flatten\{\bm{P^{3D\_com}}\}]).
\label{eq6}
\end{equation}
Then, a cross-attention is employed to extract human-relevant features from image features, which utilizes the motion context as the query $\bm{Q_m}$ and the image features as the key $\bm{K_i}$ and value $\bm{V_i}$. 
The fused features $\bm{F'}$ are used to predict SMPL shape $\bm{\beta}$ and initial pose parameters $\bm{\theta_{init}}$ for each frame. The detailed formula is as follows:
\begin{equation}
  \bm{F'} = \mathrm{MCA}(\bm{Q_m}, \bm{K_i}, \bm{V_i}),
\label{eq6}
\end{equation}
\begin{equation}
 \bm{\beta}, \bm{\theta_{init}} = \mathrm{MLP}(\bm{F'}),
\label{eq6}
\end{equation}
where $\mathrm{MCA}$ represents the multi-head cross-attention. 
This fusion effectively leverages a reliable motion prior to extract human-related information from images (e.g., body shape and initial pose). 

\noindent\textbf{Motion-aware Refinement.}
The SMPL pose uses a 6D representation~\cite{6d}, which makes it difficult to predict accurate rotation from highly extracted motion context and image features~\cite{shapeboost}. 
However, predicting SMPL poses based on joint positions through inverse kinematics~\cite{niki, hybrik} tends to be more accurate under occlusions. 
Therefore, we first decompose the SMPL pose into the swing between neighboring joints and the twist represents the rotation angle along the bone direction. 
Swing can be accurately derived from joint positions, while twist cannot be inferred from joints, so we regress each joint's twist from the fused features. 
This decomposition allows us to address the challenging SMPL pose estimation by separately calculating joint-based swing and image-based twist. The formulas are as follows:
\begin{equation}
\bm{swing} = \mathrm{MLP}(\bm{P^{3D\_com}}), \bm{twist} = \mathrm{MLP}(\bm{F'}), 
\label{eq1}
\end{equation}
\begin{equation}
\bm{{\theta}_{refined}} = \mathrm{MLP}(concat[\bm{swing}, \bm{twist}]) + \bm{{\theta}_{init}}. 
\label{eq1}
\end{equation}
Once the refined SMPL pose and shape parameters are obtained, they are fed into the pre-trained SMPL regressor~\cite{spin} to generate the final human mesh sequence $\bm{M}$.

\subsection{Loss Function}
To accelerate the convergence of the entire network, the occluded parts complement module is pre-trained on the AMASS dataset~\cite{amass}. 
The $\mathcal{L}1$ 3D joint loss is used to supervise predicted 3D joints, which is calculated as follows:
\begin{equation}
\mathcal{L}_{Pred} = \sum_{t=1}^{T}\left \|  \bm{P^{3D\_com}}-\hat{\bm{P}}^{\bm{3D}} \right \| _{1},
\label{eq1}
\end{equation}
where $\hat{\bm{P}}^{\bm{3D}}$ is the ground-truth human pose. For the end-to-end training of the entire network, consistent with previous methods~\cite{dpmesh, 3dcrowdnet}, we adopt $\mathcal{L}2$ SMPL pose and shape loss $\mathcal{L}_{pose}$, $\mathcal{L}_{shape}$ to supervise the estimation of human model parameters. 
The mesh loss $\mathcal{L}_{mesh}$ is employed to calculate the $\mathcal{L}1$ loss between the predicted 3D mesh vertices and the ground truth vertices. 
Moreover, we use 3D joint loss $\mathcal{L}_{joint}$ to further ensure the accuracy of 3D joints regressed from the mesh. 
The formulas are as follows: 
\begin{equation}
\mathcal{L}_{pose} = \left \|  \bm{\theta_{refined}}-\bm{\hat{\theta}} \right \| _{2},
\label{eq5}
\end{equation}
\begin{equation}
\mathcal{L}_{shape} = \left \|  \bm{\beta}-\bm{\hat{\beta}} \right \| _{2},
\label{eq4}
\end{equation}
\begin{equation}
\mathcal{L}_{mesh} = \left \|  \bm{M}-\bm{\hat{M}} \right \| _{1},
\label{eq2}
\end{equation}
\begin{equation}
\mathcal{L}_{joint} = \left \|  \bm{WM}-\bm{\hat{J^{3D}}} \right \| _{1},
\label{eq3}
\end{equation}
where $\bm{\hat{P^{3D}}}$, $\bm{\hat{\theta}}$, $\bm{\hat{\beta}}$, $\bm{\hat{M}}$, and $\bm{\hat{J^{3D}}}$ represent the ground truth annotations of 3D poses, SMPL pose parameters, SMPL shape parameters and 3D joints regressed from mesh, respectively.
$\bm{W}\mathbb{\in R} ^ {K\times 6890}$ is the regression matrix that can map SMPL vertices into specific human joints format. 
For training on 2D datasets\cite{instavariety}, we use the projected 2D keypoints to calculate the joint loss.
The total loss is calculated as:
\begin{equation}
\mathcal L = \lambda_{p} \mathcal L_{pose} + \lambda_{s} \mathcal L_{shape} + \lambda_{m} \mathcal L_{mesh} + \lambda_{j} \mathcal L_{joint},
\label{eq1}
\end{equation}
where $\lambda_{p}=1.0$, and $\lambda_{s}=0.001$, $\lambda_{m}=1.0$, $\lambda_{j}=5.0$ in MoPO. 
These parameters ensure that the values of different loss functions are maintained at the same range. 

\section{Experiments}
\subsection{Implementation Details}
Our MoPO is trained in two stages: 1) pre-training for the motion de-occlusion and 2) fine-tuning for the entire network. 
In the pre-training stage, the lifting model and occluded parts complement are trained on the synthetic AMASS~\cite{amass} dataset. 
AMASS provides SMPL annotations, allowing us to obtain 3D joints through joint regressor. 
To generate 2D joints input, we follow WHAM~\cite{wham} by creating a virtual camera that projects 3D joints into 2D joints. 
This stage is trained for $100$ epochs with a learning rate of $5~\times~10^{-4}$, using $K=17$ joints follow Human3.6M~\cite{h36m}. The motion predictor takes in $N=16$ frames and outputs $L=8$ frames.

\noindent\textbf{Implementations of Lifting Model.}
Different from existing lifting-based 3D pose estimation methods~\cite{hot, motionbert} that rely on complex spatial-temporal Transformer networks, our lifting model utilizes a lightweight yet effective RNN. It effectively generates accurate and smooth 3D poses based on 2D poses. 
Specifically, After obtaining 2D poses $\bm{P^{2D}}\mathbb{\in R}^{T\times K\times 2}$ from existing 2D pose detectors~\cite{vitpose, distribution}, joint coordinates are first projected to an embedding dimension through linear layer.
Then, the LSTM-based lifting model estimates the 3D poses sequentially in a frame-by-frame way. 
For each time step, the LSTM takes in the current frame’s 2D pose features, the previous frame’s 3D pose sequence, and the hidden state. Then, it estimates the 3D pose and the hidden state of the current frame.
The pose sequence records the sequence of estimated 3D poses, while the hidden state captures the motion tendency. 
Once all time steps are processed, the estimated\_list is concatenated to form the estimated 3D pose sequence $\bm{P^{3D}}\mathbb{\in R}^{T\times K\times 3}$.


\noindent\textbf{Implementations of MLP-Mixer.}
The MLP-Mixer~\cite{mlpmixer} we used only contains fully connected layers and layer normalization. 
Given the 3D poses after the Discrete Cosine Transform (DCT)~\cite{dct}, we first apply a fully connected layer to extract information on the spatial dimensions:
\begin{equation}
\bm{z^{0}} = \mathrm{DCT}(\bm{\{P^{3D}_t\}}^N_{t=0}) \bm{W_0} + \bm{b_0},
\label{eq3}
\end{equation}
where $N$ is the input length of 3D poses, $\bm{z_0}\mathbb{\in R}^{N\times C}$ is the output of the fully connected layer, $\bm{W_0}\mathbb{\in R}^{C\times C}$ and $\bm{b_0}\mathbb{\in R}^{C}$ represent the learnable parameters of the fully connected layer, $C=3\times K$ denotes the flatten keypoint coordinates.
Next, we employ a series of $m=48$ blocks MLP to extract temporal information across frames. 
Each block consists of a fully connected layer followed by layer normalization, defined by the following equation: 
\begin{equation}
\bm{z^{i}} = \bm{z^{i-1}} + \mathrm{LN}(\bm{W_i} \bm{z^{i-1}} + \bm{b_i}),
\label{eq4}
\end{equation}
where $\bm{z_i}\mathbb{\in R}^{N\times C}$, $i\mathbb{\in }[1,...,m]$ is the output of the $i$-th fully connected layer, $\mathrm{LN}$ is Layer Normalization, $\bm{W_i}\mathbb{\in R}^{C\times C}$ and $\bm{b_i}\mathbb{\in R}^{C}$ represent the learnable parameters of the $i$-th fully connected layer.
Finally, we add another fully connected layer after the MLP blocks, which also extract features on the spatial dimension. The IDCT transformation is then applied to obtain the final predictions:
\begin{equation}
\bm{\{P^{3D\_pred}_t\}}^{N+L}_{t=N} = \mathrm{IDCT}(\bm{z^{m}} \bm{W_{m+1}} + \bm{b_{m+1}}),
\label{eq4}
\end{equation}
where $\bm{P^{3D\_pred}}$ represents the predicted pose sequence, which is used to complete occluded body parts. $L$ is the predicted length of 3D poses. 

\noindent\textbf{Training Datasets.} Following previous methods~\cite{dpmesh, 3dcrowdnet}, we fine-tune the entire network end-to-end using a mix of 2D and 3D datasets: Human3.6M~\cite{h36m}, MPI-INF-3DHP~\cite{mpii3d}, MuCo-3DHP~\cite{3dhp}, and InstaVariety~\cite{instavariety}. 
{Human3.6M}~\cite{h36m} is collected in an indoor environment with four views.
The training set typically includes images from five actors (S1, S5, S6, S7, S8) with a total of 1,559,752 frames, while the test set consists of two actors (S9 and S11) with 550,644 frames for evaluation. 
The dataset provides pose annotations for 32 body joints, of which 17 joints are commonly used in research. 
{MPI-INF-3DHP}~\cite{mpii3d} is a single-person dataset collected with multiple cameras, offering greater diversity in terms of pose, appearance, clothing, and occlusions. 
It includes both indoor and outdoor scenes. We use all training subjects from S1 to S8, totaling 90K images.
{InstaVariety}~\cite{instavariety} is a large-scale in-the-wild video dataset with significant diversity in subjects, motions, and environments. We train our method on the training split of this dataset with projected 2D joints.
{MuCo-3DHP}~\cite{3dhp} is a synthetic multiperson composited dataset based on the person segmentation masks from MPI-INF-3DHP. 
The synthesis process generated realistic images covering a range of overlapping and interactive scenarios between simulated humans.

This stage is trained for $100$ epochs with a learning rate of $1~\times~10^{-4}$. 
The video clip length is set to $T=81$ following~\cite{wham}. The spatial-temporal weight $\alpha$ is set to $0.8$ and the occlusion detection thredshold $thred$ is set to 0.6.
The image features channel is $C_1=1024$ and the hidden channel of cross-attention is set to $C_2=512$. 
To prevent unrealistic joint rotations, the motion-aware fusion and refinement predicts human poses in the 6D representation~\cite{6d}. 
Finally, the SMPL pose and shape parameters are fed into the SMPL regressor from SPIN~\cite{spin} to render the final human mesh. 
All experiments are conducted using PyTorch on a single NVIDIA RTX 4090 GPU.
\subsection{Datasets and Metrics}
\textbf{Datasets.} 
To compare with previous occluded HMR methods~\cite{dpmesh, instancehmr, jotr, 3dcrowdnet}, we evaluate the performance of MoPO on the occlusion-specific benchmarks: 3DPW-OC~\cite{3doh}, 3DPW-PC~\cite{romp}, 3DPW-Crowd~\cite{3dcrowdnet}, 3DOH~\cite{3doh}, and CMU-Panoptic~\cite{cmu}. 
For the standard benchmark, we use the 3DPW~\cite{3dpw} to evaluate MoPO's robustness in wild scenes.

\noindent\textbf{Evaluation Metrics.} To evaluate the accuracy of reconstructed human mesh, we employ the mean per joint position error (MPJPE), Procrustes-aligned MPJPE (PA-MPJPE) for the evaluation of 3D joints accuracy and mean per vertex position error (MPVPE) for the evaluation of 3D mesh accuracy. 
These metrics measure the difference between the predicted positions and ground truth in millimeters ($mm$). 

\begin{table}[t]
	\centering
    \Huge
 \caption{ \textbf{Quantitative comparisons on occlusion-specific benchmarks.} Our MoPO demonstrates outstanding estimation accuracy under object-person and person-person occlusions, showing the effectiveness of our approach. All metrics are measured in millimeters $(mm)$.}
\resizebox{\textwidth}{!}
	{\begin{tabular}{lccc|ccc|ccc|ccc} 
	\toprule 
	\multirow{2}{*}{Method}& \multicolumn{3}{c}{3DPW-OC} &\multicolumn{3}{c}{3DPW-PC} &\multicolumn{3}{c}{3DOH}
 &\multicolumn{3}{c}{3DPW-Crowd}\\
 \cmidrule(lr){2-4}\cmidrule(lr){5-7}\cmidrule(lr){8-10}\cmidrule(lr){11-13}
	&MPJPE$\downarrow$ &PA-MPJPE$\downarrow$ &MPVPE$\downarrow$   &MPJPE$\downarrow$ &PA-MPJPE$\downarrow$ &MPVPE$\downarrow$    &MPJPE$\downarrow$ &PA-MPJPE$\downarrow$ &MPVPE$\downarrow$ 
 &MPJPE$\downarrow$ &PA-MPJPE$\downarrow$ &MPVPE$\downarrow$\\
    \midrule
SPIN (CVPR'19) \cite{spin}  & 95.5 & 60.7 & 121.4  & 122.1     & 77.5     & 159.8    & 110.5   & 71.6    & 124.2 & 121.2 & 69.9 & 144.1   
\\
PyMAF (ICCV'21) \cite{pymaf}  & 89.6 & 59.1   & 113.7    & 117.5 & 74.5  & 154.6 & 101.6  & 67.7  & 116.6 & 115.7 & 66.4 & 147.5
\\
ROMP (ICCV'21) \cite{romp} & 91.0  & 62.0  & -      & 98.7 & 69.0  & - & - & -  & - & 104.8 & 63.9 & 127.8
\\
OCHMR (CVPR'22)  \cite{ochmr} & 112.2  & 75.2 & 145.9  & -  & - & -   & - & -  & -  & - & - & -
\\ 
PARE (ICCV'21) \cite{pare} & 83.5 &  57.0   & 101.5   & 95.8 & 64.5  & 122.4     & 109.0 & 63.8  & 117.4    & 94.9 & 57.5 & 117.6
\\
3DCrowdNet (CVPR'22) \cite{3dcrowdnet} & 83.5 & 57.1   & 101.5   & 90.9 & 64.4   & 114.8   & 102.8 & 61.6  & 111.8  &  85.8 & 55.8 & 108.5
\\
JOTR (ICCV'23) \cite{jotr}  & 75.7  & 52.2  & 92.6   & 86.5  & 58.3 & 109.7  & 98.7  & 59.3  & \underline{104.8}  & 82.4 & 52.0 & 103.4
\\
InstanceHMR (CVPR'24) \cite{instancehmr}  & -  & -  & -   & 99.9  & 76.3  & 126.4  & -  & -  & -  & - & - & -
\\
DPMesh (CVPR'24) \cite{dpmesh} & \underline{70.9}  & \underline{48.0}  & \underline{88.0}   & \underline{82.2}  & \underline{56.6}  & \underline{105.4}  & \underline{97.1} & \underline{59.0}   &  {106.4} 
& \textbf{79.9} & \underline{51.1} & \underline{101.5}
\\
PostoMETRO (WACV'25) \cite{posto} & 79.7 & 49.0 & 99.9 & 95.3 & 61,0 & 110.9 & - & - & - & - & - & - \\
    
    \midrule
\rowcolor{cyan!10} MoPO (Ours)  &  \textbf{64.9} & \textbf{41.2} & \textbf{77.4}  & \textbf{76.0} &  \textbf{52.1}     & \textbf{101.8} & \textbf{95.7}  & \textbf{57.0} & \textbf{104.9}  & \underline{82.1}  & \textbf{45.4} &  \textbf{101.0}\\
	
	\bottomrule 
	\end{tabular}}
    
    \label{tab1}
\end{table}

\subsection{Comparisons on Occlusion Benchmarks}
To evaluate MoPO's robustness under occlusions, we compare MoPO with previous state-of-the-art methods under object-person and person-person occlusions. For a fair comparison, no method is fine-tuned on the evaluation datasets.

\noindent\textbf{Object-person Occlusion.} We evaluate MoPO’s robustness to object occlusions on the 3DPW-OC~\cite{3dpw} and 3DOH~\cite{3doh} test set. 
3DPW-OC is an object-occluded subset of 3DPW, containing 20529 images. 
As shown in Table~\ref{tab1}, MoPO outperforms DPMesh~\cite{dpmesh} across all metrics, significantly reducing MPJPE and MPVPE by 8.5\% (from 70.9 mm to 64.9 mm) and 12.0\% (from 88.0 mm to 77.4 mm), respectively. 
These improvements demonstrate that MoPO effectively handles object occlusions in the wild by accurately completing the occluded human parts. 
The 3DOH test set includes 1,290 images. Since the 3DOH test set is not a sequential dataset, we repeat the same frame $T$ times to apply MoPO. 
Despite this limitation, MoPO achieves the best results with 95.7 mm on MPJPE and 104.9 mm on MPVPE, demonstrating that the motion-aware fusion and refinement effectively leverages visible poses to improve accuracy.

\begin{table}[t]
\centering
\caption{ \textbf{Quantitative comparison on CMU-Panoptic~\cite{cmu}}. We evaluate the MPJPE for four scenes and calculate the mean performance. MoPO shows superior robustness for multi-person scenes.}
\scriptsize
\adjustbox{width=0.8\linewidth}{
      \begin{tabular}{lcccc|c}
      \toprule
        Method    & Haggl.  & Mafia & Ultim. & Pizza & Mean\\
        \midrule
        {Zanfir \textit{et al.}} (NeurIPS'18)~\cite{zanfir}  &141.4 & 152.3 & 145.0 & 162.5 & 150.3\\
        {Jiang \textit{et al.}} (CVPR'20)~\cite{jiang} & 129.6 & 133.5 & 153.0 & 156.7 & 143.2 \\
        {ROMP} (ICCV'21)~\cite{romp} & 111.8 & 129.0 & 148.5 & 149.1 & 134.6 \\
        {REMIPS} (NeurIPS'21)~\cite{remips} & 121.6 & 137.1 & 146.4 & 148.0 & 138.3\\
        {3DCrowdNet}~(CVPR'22)~\cite{3dcrowdnet} & 109.6 & 135.9 & 129.8 & 135.6 & 127.6\\
        {JOTR}~(ICCV'23)~\cite{jotr} & 99.9 & 113.5 & 115.7 & 123.6 & 114.7\\
        DPMesh~(CVPR'24)~\cite{dpmesh} & {97.2} & {109.8} & {114.3} & {120.5} & {110.4}\\
        \midrule
        \rowcolor{cyan!10} MoPO (Ours) & \textbf{95.3} & \textbf{107.1} & \textbf{113.7} & \textbf{117.4} & \textbf{108.4}\\
        
        \bottomrule
    \end{tabular}
  }
  
  \label{tab3}
\end{table}
\noindent\textbf{Person-person Occlusion.} 
To evaluate the robustness of MoPO under person occlusions, we use 3DPW-PC~\cite{romp}, 3DPW-Crowd~\cite{3dcrowdnet}, and CMU-Panoptic~\cite{cmu} for testing. 
Person-person occlusion not only causes joint occlusions but also introduces uncertain joint-human ownership due to overlapping limbs, impacting the accuracy of 2D pose estimation and 3D pose lifting modules~\cite{instancehmr}. 
3DPW-PC is a person-occluded subset of 3DPW, including 1380 images and 2,218 persons. 
As shown in Table~\ref{tab1}, MoPO surpasses previous methods on 3DPW-PC with 76.0 mm MPJPE and 101.8 mm MPVPE, indicating that motion prior can help alleviate joint-human ambiguity. 
3DPW-Crowd is the person crowded subset of 3DPW, containing 1073 images and 1,923 persons. MoPO outperforms previous methods in PA-MPJPE and MPVPE, achieving 45.4 mm and 101.0 mm. 
However, MoPO's MPJPE is slightly higher than DPMesh~\cite{dpmesh} due to the inaccurate 2D poses in almost all frames, which limits the reliability of motion prior under extremely crowded scenarios.
The CMU-Panoptic dataset is a multi-view, multi-person indoor dataset. Following~\cite{dpmesh, 3dcrowdnet}, we choose four scenes (haggl, mafia, ultim, pizza) for testing, including 9,600 images and 21,404 persons. As shown in Table~\ref{tab3}, MoPO surpasses existing methods in MPJPE across all scenes, yielding an average improvement of 2.0 mm over DPMesh.

\begin{table}[t]
\centering
\small
\caption{ \textbf{Quantitative comparisons on 3DPW~\cite{3dpw} test split.} All methods are tested without training on the 3DPW training set. MoPO obtains significantly higher pose and shape accuracy.}
\adjustbox{width=0.7\linewidth}
  {
      \begin{tabular}{lcccc}
      \toprule
        \multirow{2}{*}{Method} & \multicolumn{3}{c}{3DPW} \\
        \cmidrule(lr){2-4}
        & MPJPE$\downarrow$  & PA-MPJPE$\downarrow$ & MPVPE$\downarrow$\\
        \midrule
        {HMR}~(CVPR'18)~\cite{hmr}  &130.0 &76.7 & - \\
        {PyMaf}~(ICCV'21)~\cite{pymaf} & 92.8 & 58.9 & 110.1 \\
        {PARE}~(ICCV'21)~\cite{pare} & 82.9 & 52.3 & 99.7 \\
        {GLAMR}~(CVPR'22)~\cite{glamr} & - & 51.1 & - \\
        {3DCrowdNet}~(CVPR'22)~\cite{3dcrowdnet} & 81.7 & 51.2 & 98.3 \\
        {SEFD}~(ICCV'23)~\cite{sefd} &77.4 &49.4 & 92.6 \\
        DPMesh (CVPR'24)\cite{dpmesh} & 73.6 & 47.4 & 90.7 \\
        InstanceHMR (CVPR'24)\cite{instancehmr} & 73.2 & 44.3 & 80.3 \\
        GraphMLP~(PR'25)~\cite{graphmlp} &140.5 & 66.7 & - \\
        ASFnet~(PR'26)~\cite{asfnet} &108.8 & 60.1 & - \\
        PostoMETRO (WACV'25)\cite{posto} & 84.9 & 48.9 & - \\
        MEGA (CVPR'25)\cite{mega} & \underline{67.5} & \underline{41.0} & \underline{80.0} \\
        \midrule
        \rowcolor{cyan!10}MoPO (Ours) & \textbf{60.6} & \textbf{37.1} & \textbf{71.2} \\
        
        \bottomrule
      \end{tabular}
  }
  \label{tab2}
\end{table}
\subsection{Comparisons on Standard Benchmark}
\textbf{Comparisons with Occlusion Specific Models.} 3DPW is a widely used in-the-wild dataset for human mesh recovery task and contains well-processed 3D joint and SMPL annotations. 
As shown in Table~\ref{tab2}, compared to the state-of-the-art method MEGA~\cite{mega}, MoPO achieves a 10.2\% reduction in MPJPE (from 67.5 $mm$ to 60.6 $mm$), a 9.5\% decrease in PA-MPJPE (from 41.0 $mm$ to 37.1 $mm$), and a significant 11.0\% improvement in MPVPE (from 80.0 $mm$ to 71.2 $mm$).
This is because the motion prior in less occluded scenarios is more reliable and sufficient for HMR.
Moreover, these results demonstrate MoPO's superior performance on general in-the-wild data without overfitting to occlusion-specific datasets.

\begin{table}[t]
\centering
\small
\caption{ \textbf{Comparisons with image-based and video-based standard models.} MoPO achieves state-of-the-art among general non occlusion specific methods. ``*" represents using 3DPW for training.}
\adjustbox{width=0.8\linewidth}
  {
      \begin{tabular}{llcccc}
      \toprule
        \multirow{2}{*}{Method} & & \multicolumn{3}{c}{3DPW} \\
        \cmidrule(lr){3-5}
        & & MPJPE$\downarrow$  & PA-MPJPE$\downarrow$ & MPVPE$\downarrow$\\
        \midrule
        \multirow{5}{*}{\shortstack{Image-\\based}}
        &{MeshPose}*~(CVPR'24)~\cite{meshpose} & 76.1 & 46.7 & 92.7 \\
        &{ShapeBoost}*~(AAAI'24)~\cite{shapeboost} & 75.3 & 44.6 & - \\
        &{TokenHMR}~(CVPR'24)~\cite{tokenhmr} & 71.0 & 44.3 & 84.6 \\
        &{PostureHMR}~(CVPR'24)~\cite{posturehmr} & 64.9 & 39.6 & 75.4 \\
        &{ScoreHypo}*~(CVPR'24)~\cite{scorehypo} & {61.8} & \textbf{36.1} & \underline{71.9} \\
        \midrule
        \multirow{5}{*}{\shortstack{Video-\\based}}
        &{MEAD}~(ICCV'21)~\cite{mead} & 79.1 & 45.7 & 92.6 \\
        &{MPS-Net}~(CVPR'22)~\cite{mpsnet} & 84.3 & 52.1 & 99.7 \\
        &{GLoT}*~(CVPR'23)~\cite{glot} & 80.7 & 50.6 & 96.3 \\
        &{PMCE}*~(CVPR'23)~\cite{pmce} & 69.5 & 46.7 & 84.8 \\
        &{ARTS}*~(MM'24)~\cite{arts} & 67.7 & 46.5 & 81.4 \\
        &{WHAM}~(CVPR'24)~\cite{wham} & \underline{60.8} & {37.8} & {72.5} \\
        \midrule
        \multirow{3}{*}{\shortstack{Occlusion \\ video}}
        & {CHOMP}~(arXiv'22)~\cite{chomp} & 83.7 & 51.7 & 110.1 \\
        & {GLMAR}~(CVPR'22)~\cite{glamr} & - & 51.1 & - \\ 
        & \cellcolor{cyan!10}MoPO (Ours) & \cellcolor{cyan!10}\textbf{60.6} & \cellcolor{cyan!10}\underline{37.1} & \cellcolor{cyan!10}\textbf{71.2} \\
        \bottomrule
      \end{tabular}
  }
  \label{tabb}
\end{table}
\noindent\textbf{Comparisons with Image-based Models.}
We further compared MoPO with image-based Human Mesh Recovery (HMR) methods under general scenes. 
As shown in Table~\ref{tabb}, MoPO outperforms the latest SOTA methods in both MPJPE and MPVPE, highlighting that occlusion is a significant limitation for existing approaches and effectively handling occlusions can substantially improve reconstruction accuracy. 
Although ScoreHypo~\cite{scorehypo} achieves a slightly lower PA-MPJPE (1 mm less than MoPO), this improvement is attributed to its use of 3DPW as part of its training dataset, which provides an advantage in performance.

\noindent\textbf{Comparisons with Video-based Models.}
As shown in Table~\ref{tabb}, we further compare MoPO with other methods that leverage temporal information. 
Currently, few methods~\cite{chomp, glamr} incorporate temporal information for occlusion handling Human Mesh Recovery (HMR), since some large-scale training datasets consist of discontinuous frames. However, our approach uses sequential datasets for training. 
Compared to GLAMR~\cite{glamr}, which infills occluded human parts by leveraging a learned implicit latent code of human dynamics, our explicit motion de-occlusion is more reliable and achieves better performance. 
Additionally, unlike existing video-based methods~\cite{mead, mpsnet, glot, pmce, arts, wham}, our MoPO achieves state-of-the-art performance on 3DPW by effectively handling occlusions, as video-based methods often produce unstable predictions under occlusions.

\subsection{Ablation Study}

\textbf{Effectiveness of Spatial-Temporal Occlusion Detector.} 
Table~\ref{tab4} (row 1-2) compares our spatial-temporal occlusion detector with other implementations. 
Using only spatial confidence like previous methods~\cite{dpmesh, 3dcrowdnet} results in a 6.5 mm decline in MPJPE on the 3DPW, which {highlights the significant role of temporal information in enhancing detection consistency}.
Furthermore, relying solely on temporal information for detection (i.e., using only RNN-estimated confidence) leads to all metrics decrease. This decline occurs because the lack of spatial visibility from the current frame hampers accurate occlusion detection.

\begin{table}[t]
	\centering
    \Huge
 \caption{ \textbf{Ablation studies.} We conduct experiments on 3DPW~\cite{3dpw} and 3DPW-OC~\cite{3doh} to validate the effectiveness of different modules and different implementations within MoPO. 
    }
\large
 \adjustbox{width=0.7\linewidth}
	{\begin{tabular}{lcc|cc} 
	\toprule 
	\multirow{2}{*}{Settings}& \multicolumn{2}{c}{3DPW} &\multicolumn{2}{c}{3DPW-OC}\\
 \cmidrule(lr){2-3}\cmidrule(lr){4-5}
	&MPJPE$\downarrow$ &MPVPE$\downarrow$  &MPJPE$\downarrow$ &MPVPE$\downarrow$ \\
    \midrule
    
    \multicolumn{3}{l}{\textbf{\textit{Occlusion Detection}}} \\
    (1) spatial only&67.1 & 80.5 &81.1 &98.1 \\
    (2) temporal only&69.4 & 81.1 &84.5 &96.8 \\
    \rowcolor{cyan!10}{(3) spatial-temporal detector}&\textbf{60.6}& \textbf{71.2}& \textbf{64.9} & \textbf{77.4}\\
    \midrule
    
    \multicolumn{3}{l}{\textbf{\textit{Occluded Parts Complement}}} \\
        (4) estimated poses&75.8 & 94.4 &76.8 &92.0 \\
        (5) estimated poses w. conf &70.9 & 83.8 &73.6 &90.7 \\
       \rowcolor{cyan!10}(6) completed poses &\textbf{60.6}& \textbf{71.2}& \textbf{64.9} & \textbf{77.4}\\
    \midrule
    \multicolumn{4}{l}{\textbf{\textit{Type of Feature Fusion}}} \\
    (7) image features only &67.1 & 78.9 &69.4 &84.5 \\
    (8) motion context only &63.6 & 75.1 &65.8& 79.2 \\
    (9) MLP-based fusion &62.0 & 73.9 &65.5 &78.9 \\
    \rowcolor{cyan!10}(10) cross-attention&\textbf{60.6}& \textbf{71.2}& \textbf{64.9} & \textbf{77.4}\\

    \multicolumn{4}{l}{\textbf{\textit{Type of Motion Refinement}}} \\
    (11) w/o refinement &62.7 & 72.9 &65.8 &79.0 \\
    (12) MLP-based refinement &61.7 & 71.5 &65.7 &78.1 \\
    \rowcolor{cyan!10}(13) swing-twist decomposition&\textbf{60.6}& \textbf{71.2}& \textbf{64.9} & \textbf{77.4}\\


    \multicolumn{4}{l}{\textbf{\textit{Input and Prediction Length}}} \\
    (14) 4: 2 &65.8 & 77.0 &69.7 &82.0 \\
    (15) 8: 4 &63.0 & 72.8 &67.3 &79.4 \\
    (16) 16: 1 &64.9 & 77.8 &71.4 &81.5 \\
    \rowcolor{cyan!10} (17) 16: 8 &\textbf{60.6}& \textbf{71.2}& \textbf{64.9} & \textbf{77.4}\\

	\bottomrule 
	\end{tabular}
 }   
    \label{tab4}
\end{table}
\noindent\textbf{Effectiveness of Occluded Parts Complement.} 
Our method is the first to leverage motion predictor for completing occluded body parts. 
To investigate the effectiveness of this module, we directly use estimated 3D poses for motion-aware fusion and refinement. 
As shown in Table~\ref{tab4} (row 4), this increases MPJPE by 11.9 mm on the 3DPW-OC. 
This is because the {2D pose detector and lifting module are not optimized for handling occlusions, causing incorrect and misdirected human pose for the following network}. 
Table~\ref{tab4} (row 5) shows that although incorporating joint confidence scores into the network helps slightly by identifying potentially occluded joints, it still fails to accurately estimate joint positions due to the lack of motion prior from other frames.

\noindent\textbf{Design of Occlusion-alleviated Fusion.} 
We also conduct experiments to analyze the impact of fusing motion context with image features. 
As shown in Table~\ref{tab4} (row 7-8), regressing SMPL parameters directly from motion context decreases MPVPE by 3.9 mm on the 3DPW due to the lack of body shape information within joints. 
Using only image features for SMPL parameter regression leads to a drop in MPJPE by 6.5 mm on the 3DPW, as {it's challenging to map low-resolution image features into SMPL rotation representation~\cite{shapeboost}}. 
As shown in Table~\ref{tab4} (row 9), when using an MLP for feature fusion, the domain gap between motion context and image features makes it difficult to fully leverage both. 
In contrast, using cross-attention to fuse motion and image features can predict more accurate SMPL pose and shape.

\noindent\textbf{Design of Motion-aware Refinement.} 
We further compare our motion-aware refinement with different implementations. 
As shown in Table~\ref{tab4} (row 11-12), omitting inverse kinematics for optimization results in higher MPJPE by 2.1 mm on 3DPW, since mapping fused features to SMPL’s axis-angle representation proves challenging~\cite{shapeboost}. 
Additionally, predicting SMPL pose directly from pose sequence through MLP reduces MPJPE by 1.1 mm on the 3DPW, as {3D joint positions lack twist information}.

\noindent\textbf{Impact of Motion Prediction Length}
The length of input and predict frames is crucial for motion de-occlusion.
We investigate the impact of different input and prediction lengths. 
Following other motion prediction methods~\cite{hmpsurvey,simlpe}, we maintain a prediction length as half of the input length.
As shown in Table~\ref{tab4} (row 14-15), {increasing the input length provides more sufficient temporal information for accurately predicting occluded joints}.
Additionally, Table~\ref{tab4} (row 16) shows that predicting only one frame (16: 1) often leads to overfitting, causing a 4.3 mm decline in MPJPE on 3DPW.

\begin{table}[t]
\centering
\small
\caption{ \textbf{Quantitative comparisons on different backbones.} MoPO obtains significantly higher pose and mesh accuracy with common used backbones (ResNet50~\cite{spin}, HRNet~\cite{cliff}, ViT~\cite{4dhmr}).}
\adjustbox{width=0.8\linewidth}
  {
      \begin{tabular}{lcccc}
      \toprule
        \multirow{2}{*}{Method}  & \multirow{2}{*}{Backbone} & \multicolumn{3}{c}{3DPW} \\
        \cmidrule(lr){3-5}
        & & MPJPE$\downarrow$  & PA-MPJPE$\downarrow$ & MPVPE$\downarrow$\\
        \midrule
        DPMesh~\cite{dpmesh} & Stable Diffusion~\cite{diffusion}  & {73.6} & {47.4} & {90.7}       \\
        InstanceHMR~\cite{instancehmr} & ResNet50~\cite{spin}  & 73.2 & 44.3 & 80.3                 \\
        \midrule
        MoPO & ResNet50~\cite{spin} & 67.1 & 42.5 & 78.9 \\
        MoPO & HRNet~\cite{cliff} & 64.7 & 40.8 & 76.7 \\
        \rowcolor{cyan!10}MoPO & ViT~\cite{4dhmr} & \textbf{60.6} & \textbf{37.1} & \textbf{71.2} \\
        
        \bottomrule
      \end{tabular}
  }
  \label{tabala}
\end{table}
\textbf{Different Backbones.}
As shown in Table~\ref{tabala}, we further explore the impact of different backbones for extracting image features, demonstrating MoPO’s effectiveness across other pre-trained backbones (ResNet50~\cite{spin}, HRNet~\cite{cliff}). 
For most experiments, we used ViT~\cite{4dhmr} as the backbone, which is significantly lighter than DPMesh's stable diffusion backbone. 
When we use the same backbone with InstanceHMR~\cite{instancehmr}, MoPO still surpasses it in all metrics.
Moreover, as MoPO relies more on the completed motion prior rather than image features, where image features supplement body shape information to improve MPVPE metric.

\begin{table}[t]
\centering
\caption{ \textbf{Computational efficiency.} Comparison of the number of parameters, FLOPs and MPJPE on the 3DPW testset.}
\scriptsize
\adjustbox{width=0.8\linewidth}{
      \begin{tabular}{lcc|c}
      \toprule
        Method    & Parameters (M)$\downarrow$  & FLOPs (G)$\downarrow$ & MPJPE$\downarrow$ \\
        \midrule
        {3DCrowdNet}~(CVPR'22)~\cite{3dcrowdnet} & 30.5 & 0.41& 81.7 \\
        {JOTR}~(ICCV'23)~\cite{jotr} & 89.2 & 2.68 & 76.4 \\
        DPMesh~(CVPR'24)~\cite{dpmesh} & 2607.4 & 168.51 & \underline{{73.6}}  \\
        \midrule
        \rowcolor{cyan!10} MoPO (Ours) & \underline{85.0} & \underline{1.48} & \textbf{60.6}  \\
        
        \bottomrule
    \end{tabular}
  }
  \label{tab5}
\end{table}
\noindent\textbf{Computational Efficiency.}
As shown in Table~\ref{tab5}, {MoPO outperforms DPMesh~\cite{dpmesh} in MPJPE on 3DPW with fewer parameters and lower FLOPs using the same GPU}. 
DPMesh typically has a significantly higher network complexity due to its large and time-consuming pre-trained diffusion model. 
Compared to 3DCrowdNet~\cite{3dcrowdnet}, our MoPO achieves a substantial 21.1mm improvement in MPJPE on the 3DPW testset with an acceptable increase in complexity. This is attributed to the lightweight MLP-Mixer based  motion predictor and the effective use of motion prior.

\begin{figure*}[h]
\centering
\includegraphics[width=0.9\textwidth]{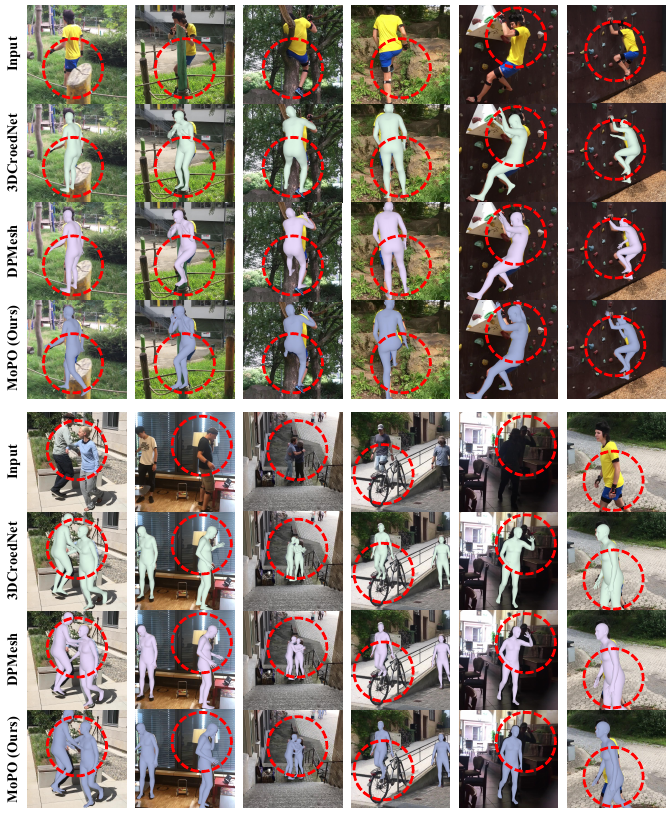}
\caption{Qualitative comparison among 3DCrowdNet \cite{3dcrowdnet} (green mesh), DPMesh \cite{dpmesh} (pink mesh), and our MoPO (blue mesh).}
\label{fig3}
\end{figure*}
\subsection{Qualitative Evaluation}
\textbf{Qualitative Comparisons on 3DPW.} 
We present qualitative comparisons among MoPO and previous SOTA methods~\cite{dpmesh, 3dcrowdnet} on the 3DPW-OC and 3DPW-Crowd dataset. 
As shown in Fig.~\ref{fig3}, MoPO achieves the best reconstruction under various occlusions, including object-person occlusions, person-person occlusions, and self-occlusions. 
Notably, under severe occlusions, previous methods often produce unnatural human pose due to the lack of human observations in image.
In contrast, MoPO leverages motion prior to generate more accurate and consistent human poses.

\begin{figure*}[tbhp]
\centering
\includegraphics[width=\textwidth]{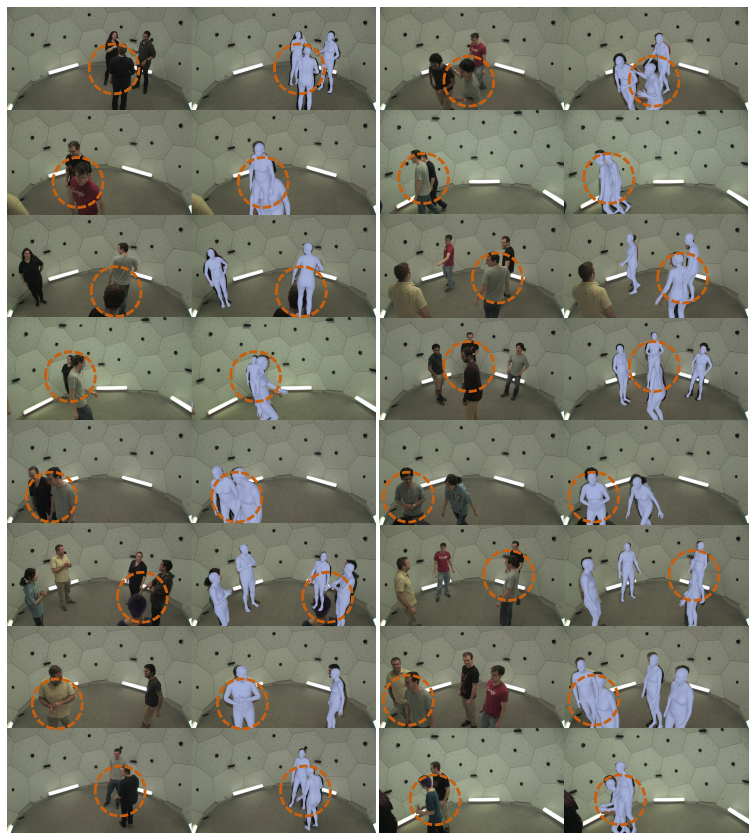}
\caption{ {Qualitative results of MoPO on CMU-Panoptic test sequences, demonstrating the robustness to multi-person scenes and diverse person occlusions.}}
\label{figb}
\end{figure*}
\clearpage
\noindent\textbf{Qualitative Results on CMU-Panoptic.}
As shown in Fig.~\ref{figb}, we show the qualitative results of MoPO on the multi-person CMU-Panoptic~\cite{cmu} test sequences. 
MoPO accurately estimates the human poses and shapes of all individuals within the view, effectively handling ambiguous person occlusions and body truncations.

\begin{figure*}[tbhp]
\centering
\includegraphics[width=\textwidth]{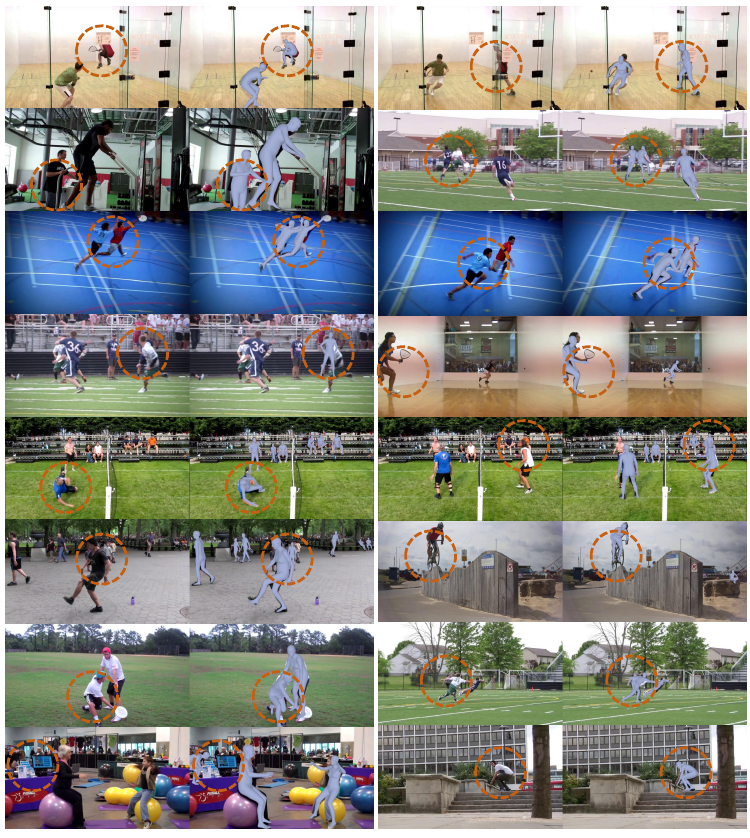}
\caption{ {Qualitative results of MoPO on the Internet videos. MoPO shows impressive generalization ability under various in-the-wild occlusions}}
\label{figc}
\end{figure*}
\noindent\textbf{Qualitative Results on the Internet Videos.}
As shown in Fig.~\ref{figc}, we evaluate the generalization ability of MoPO on various Internet videos. 
When applied to in-the-wild videos with complex backgrounds, diverse human motions, and heavy occlusions, MoPO demonstrates robust generalization. 
This highlights that the use of explicit motion prior significantly enhances the reliability and stability of HMR.

\begin{figure*}[tbhp]
\centering
\includegraphics[width=\textwidth]{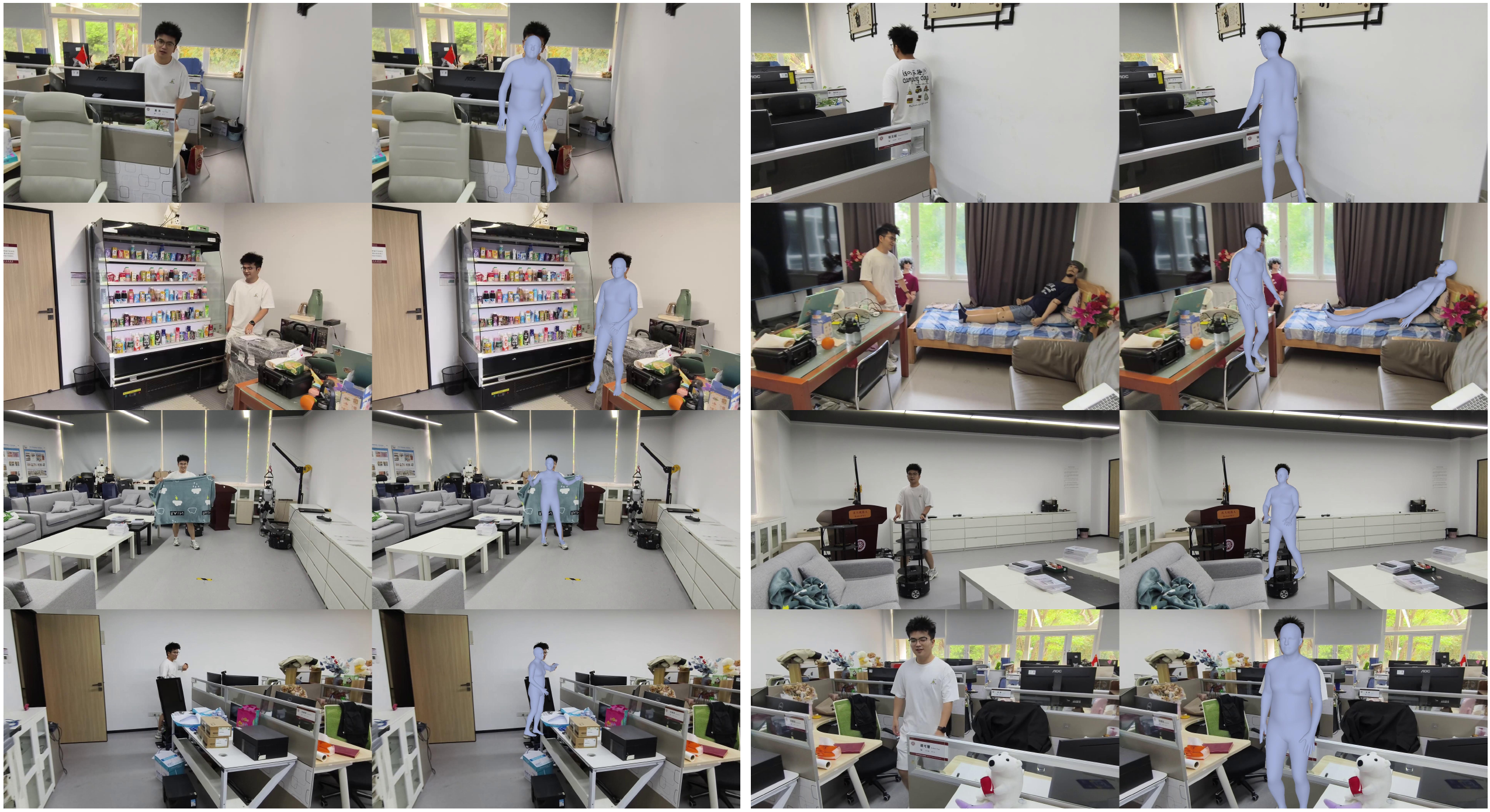}
\caption{ {Qualitative results of MoPO in real-world occlusion scenarios}}
\label{figd}
\end{figure*}

\noindent\textbf{Qualitative results of MoPO in real-world occlusion scenarios.}
As shown in Fig. \ref{figd}, MoPO demonstrates superior robustness when handling real-world occlusions in daily activities. For partial occlusion (e.g., self-occlusion or joints partially blocked by objects), MoPO effectively utilizes skeletal motion priors to physically constrain missing joints, preventing unnatural poses. For heavy occlusion (e.g., large portions of the torso or limbs are obscured), our motion-aware fusion strategy accurately infers the spatial positions of occluded parts by leveraging long-term motion trends. These resMoults verify that MoPO achieves exceptional reconstruction robustness in complex, real-world occlusion scenarios.

\begin{figure}[htbp]
    \centering    
    \includegraphics[width=0.9\linewidth]{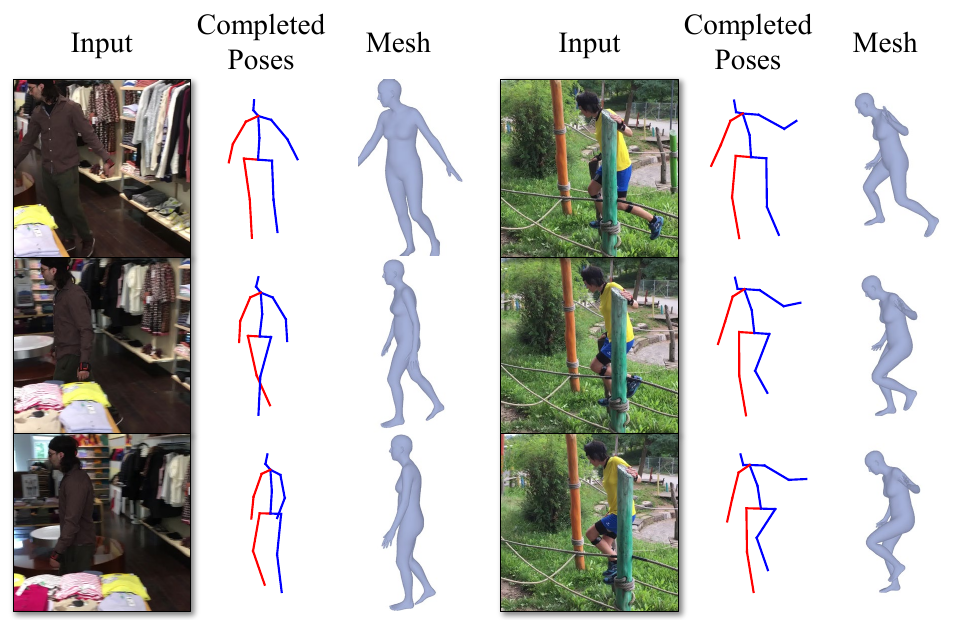}
    \caption{
        Visualization of completed poses and human mesh on the object-person occlusion sequence.
    }
    \label{fig4}
\end{figure}
\noindent\textbf{Completed Poses Visualizations.} 
To further demonstrate the role of motion de-occlusion, we visualize the completed poses for occluded frames from object-person occluded sequences from 3DPW dataset.
As illustrated in Fig.~\ref{fig4}, when a subject is partially or significantly obscured by environmental objects (e.g., tables or trees), MoPO effectively 
uses motion prior from previous frames to predict the plausible positions of occluded joints, providing accurate human poses for the following networks.

\begin{figure}[htbp]
    \centering    
    \includegraphics[width=\linewidth]{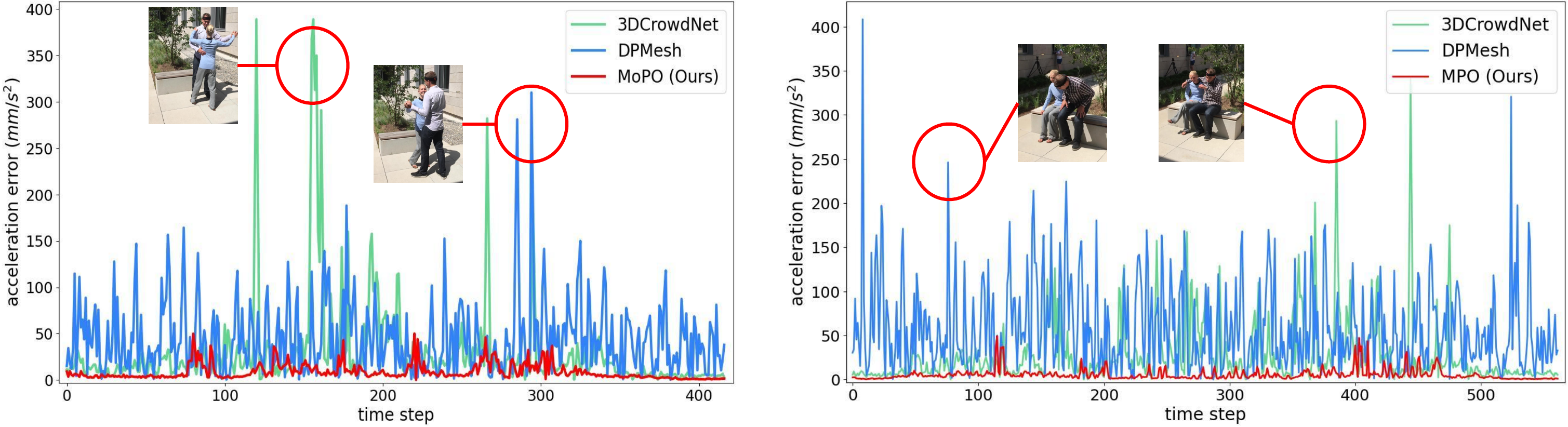}
    \caption{
        Comparison of the acceleration error for 3DCrowdNet, DPMesh, and our MoPO on the person-person occlusion sequence.
    }
    \label{fig5}
\end{figure}

\noindent\textbf{Comparison of Acceleration Error.}
Fig.~\ref{fig5} shows the acceleration error of the "courtyard\_dancing\_00" (left) and "courtyard\_hug\_00" (right) sequence in the 3DPW-Crowd dataset among MoPO and previous SOTA methods. 
Acceleration error~\cite{hmmr} indicates the temporal consistency of human motion.
When severe person occlusions occur, previous occluded HMR methods often suffer from high acceleration error and motion jitter due to the unstable estimation under occlusions. 
In contrast, MoPO maintains smooth and temporally consistent motion by incorporating motion prior, ensuring more stable motion acceleration.

\section{Conclusion}
In this paper, we propose MoPO, a novel framework that fully leverages motion prior to enhance occluded human mesh recovery from video. 
We first introduce a motion de-occlusion module, which detects occluded human joints based on spatial-temporal occlusion cues and then completes these occluded human joints through the MLP-based motion predictor. 
After obtaining the completed joint sequence, we design a motion-aware fusion and refinement module that effectively combines motion context with image features and refines plausible SMPL poses through inverse kinematics. 
Our MoPO enhances temporal consistency and mitigates unnatural poses under diverse occlusions. 
Extensive experiments demonstrate that our method achieves superior performance on both occlusion-specific and standard datasets. 
We hope that our work can inspire further research into leveraging human motion prior as a more reliable foundation for occluded human mesh recovery.
 
\section*{Acknowledgments}
This work was supported by the National Natural Science Foundation
of China (Grant No. 62373009), the Guangdong S\&T Program (Grant
No. 2024B0101050002), and the Shenzhen Innovation in Science and
Technology Foundation for The Excellent Youth Scholars (Grant No.
RCYX20231211090248064). 
\bibliographystyle{elsarticle-num}
\bibliography{mopo}

\end{document}